\documentclass[11pt,a4paper]{article}
\usepackage[hyperref]{emnlp2020}
\usepackage{times}
\usepackage{latexsym}
\usepackage{graphicx}
\usepackage{multirow}
\usepackage{booktabs}
\usepackage{comment}
\usepackage{float}
\usepackage{caption}
\usepackage{subcaption}
\usepackage{natbib}
\usepackage{amsmath, bm, mathtools}
\usepackage{CJKutf8}
\graphicspath{ {./sections/figs/} } 

\usepackage{microtype}

\aclfinalcopy 

\title{What makes multilingual BERT multilingual?}

\author{Chi-Liang Liu\thanks{$^*$Equal Contribution} \quad Tsung-Yuan Hsu$^*$ \quad Yung-Sung Chuang \quad Hung-yi Lee\\
College of Electrical Engineering and Computer Science, National Taiwan University \\
{ \tt \{liangtaiwan1230, sivia89024, tlkagkb93901106\}@gmail.com} \\
{ \tt b05901033@ntu.edu.tw}\\
}

\date{}

\begin{document}
\maketitle
\footnotetext{This work was supported by Delta Electronics, Inc..}
\begin{abstract}
Recently, multilingual BERT works remarkably well on cross-lingual transfer tasks, superior to static non-contextualized word embeddings. In this work, we provide an in-depth experimental study to supplement the existing literature of cross-lingual ability. We compare the cross-lingual ability of non-contextualized and contextualized representation model with the same data. We found that datasize and context window size are crucial factors to the transferability.
\end{abstract}

\section{Introduction}

Cross-lingual word embedding is to learn embeddings in a shared vector space for two or more languages. A line of works assumes that monolingual word embeddings share similar structures across different languages and try to impose post-hoc alignment through a mapping~\citep{mikolov:13, Smith:17, joulin:18, Lample:18, artetxe:18, Zhou:19}. Another line of works considers joint training, which optimizes monolingual objective with or without cross-lingual constraints when training word embeddings~\citep{luong:15,gouws:15,Ammar:16,duong:16,lample:18b}. Cross-lingual word embedding methods above were initially proposed for non-contextualized embedding such as GloVe~\citep{pennington:14} and Word2Vec~\citep{mikolov:13w2v}, and later adapted to contextualized word representation~\citep{schuster:19, aldarmaki-diab:19}. 

Multilingual BERT (m-BERT) \cite{devlin:19} has shown its superior ability in cross-lingual transfer on many downstream tasks, either in a way it is used as a feature extractor or finetuned end-to-end~\citep{conneau:18, wu:19, hsu-etal:19, pires:19}. 
It seems that m-BERT has successfully learned a set of cross-lingual representations in a shared vector space for multiple languages~\citep{Cao:20}. 
However, given the way how m-BERT was pre-trained, it is unclear how it succeeded in building up cross-lingual ability without parallel resources and explicit supervised objectives.

There is a line of work studying the key components contributing to the cross-lingual ability of m-BERT~\citep{Karth:20, tran:20, Cao:20, singh:19}. 
It was shown that  {\it depth} and {\it total number of parameters} remarkably affect cross-lingual ability~\citep{Cao:20}. 
The conclusion about the impact of shared vocabulary is mixed~\citep{Karth:20, singh:19}, showing that our understandings about it are still in the early stages.

In this paper, we study the impacts of some critical factors on the cross-lingual ability of m-BERT to enrich our understandings of how to build a powerful cross-lingual model.  
The contributions of this work can be summarized as the following:
\begin{itemize}
    \item We found that large enough datasizes and modeling long term dependency are all necessary factors for the cross-lingual ability of m-BERT.
    \item We found that the non-contextualized word embedding training under the same condition as m-BERT does not show the same cross-lingual ability, which shows the uniqueness of m-BERT. 
\end{itemize}



\section{How to Build up Cross-lingual Ability}
\label{section:align}

\subsection{Metrics for Cross-lingual Ability}
There are two main paradigms for evaluating cross-lingual representations:~{\it word retrieval} and {\it downstream task transfer}. 
Here we use both as the indicators of cross-lingual ability.
Although word retrieval is a task originally proposed to measure cross-lingual alignment at the word level, contextual version of word retrieval has been proposed for contextualized embeddings and consistent with downstream task transfer performance~\citep{Cao:20}. 

\subsubsection{Word Retrieval}
Given a word and a bilingual dictionary~$D=\{(x_1,y_1),(x_2,y_2),\ldots,(x_n,y_n)\}$, listing all parallel word pairs from source and target languages, word retrieval is the task to retrieve the corresponding word in target language considering information provided by embedding vectors~$\{(u_1,v_1),(u_2,v_2),...(u_n,v_n)\}$. Specifically we consider a nearest neighbor retrieval function
\begin{equation}
    \text{neighbor}(i)=\mathop{\arg\max}_{j}\text{sim}(u_i,v_j),
\end{equation}
where $u_i$ is the embedding of source word $x_i$ and we want to find its counterpart $y_i$ among all candidates~$y_1,y_2,...y_n$. 
We use cosine similarity as the similarity function $\text{sim}$.

Then we have{\it~mean reciprocal rank }(MRR) as evaluation metrics. 
\begin{equation}
\text{MRR} = \frac{1}{n}\sum_{i}^{n}\frac{1}{\text{rank}(y_i)}
\end{equation}
where $\text{rank}(y_i)$ is a ranking function based on retrieval results.
For contextualized embeddings, we average embeddings in all contexts and use the mean vector to represent each word, so that contextualized embeddings could also be evaluated with the task defined above.

\subsubsection{Downstream Task Transfer}
We consider XNLI as our downstream task to evaluate cross-lingual transfer. The XNLI dataset was constructed from the English MultiNLI dataset by keeping the original training set but human-translating development and test sets into other 14 languages~\citep{conneau:18, williams:18}. 
As there are only training data in English, models should perform zero-shot cross-lingual transfer on development and test sets.

\subsection{Experimental Setup}
To compare non-contextualized and contextualized embedding models, we conducted experiments with GloVe, Word2Vec and, BERT, where the number of dimensions was all set to 768. 
To eliminate the effect of tokenization, we use the wordpiece tokenizer to tokenized data for all the above mentioned models. 

Embeddings were first pretrained from scratch and then evaluated on word retrieval and XNLI, to assess their cross-lingual ability.   
For pretraining data, we used Wikipedia from 15 languages~(English, French, Spanish, German, Greek, Bulgarian, Russian, Turkish, Arabic, Vietnamese, Thai, Chinese, Hindi, Swahili, and Urdu) to pre-train all word embeddings following unsupervised joint training scenario, assuring each target language in the downstream task has been well pre-trained. 

For word retrieval task, we evaluated cross-lingual alignment between English and each of the remaining 14 languages, using bilingual dictionaries from MUSE\footnote{https://github.com/facebookresearch/MUSE}.
For XNLI zero-shot transfer task, the training set is in English, and the target languages of testing sets are the same as those used in the word retrieval task.  

\subsection{Datasize v.s. Model}
\begin{figure*}[t!]
\vspace{-0.5cm}
\centering
\begin{subfigure}[b]{\columnwidth}
\includegraphics[width=\linewidth]{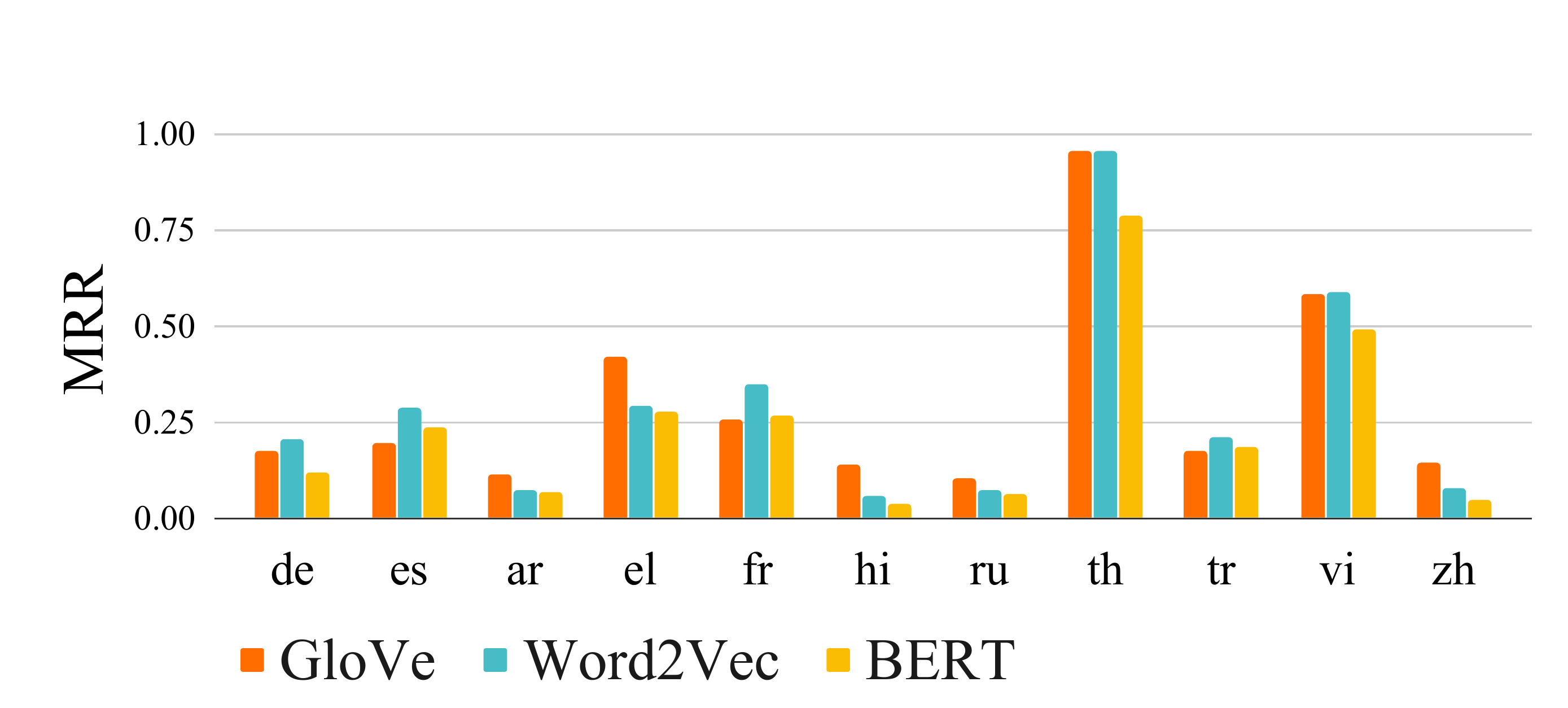}
 \subcaption{Pre-trained on 200k sentences per language.}
 \label{fig:3_1}
\end{subfigure}
\hfill
\begin{subfigure}[b]{\columnwidth}
\centering 
\includegraphics[width=\linewidth]{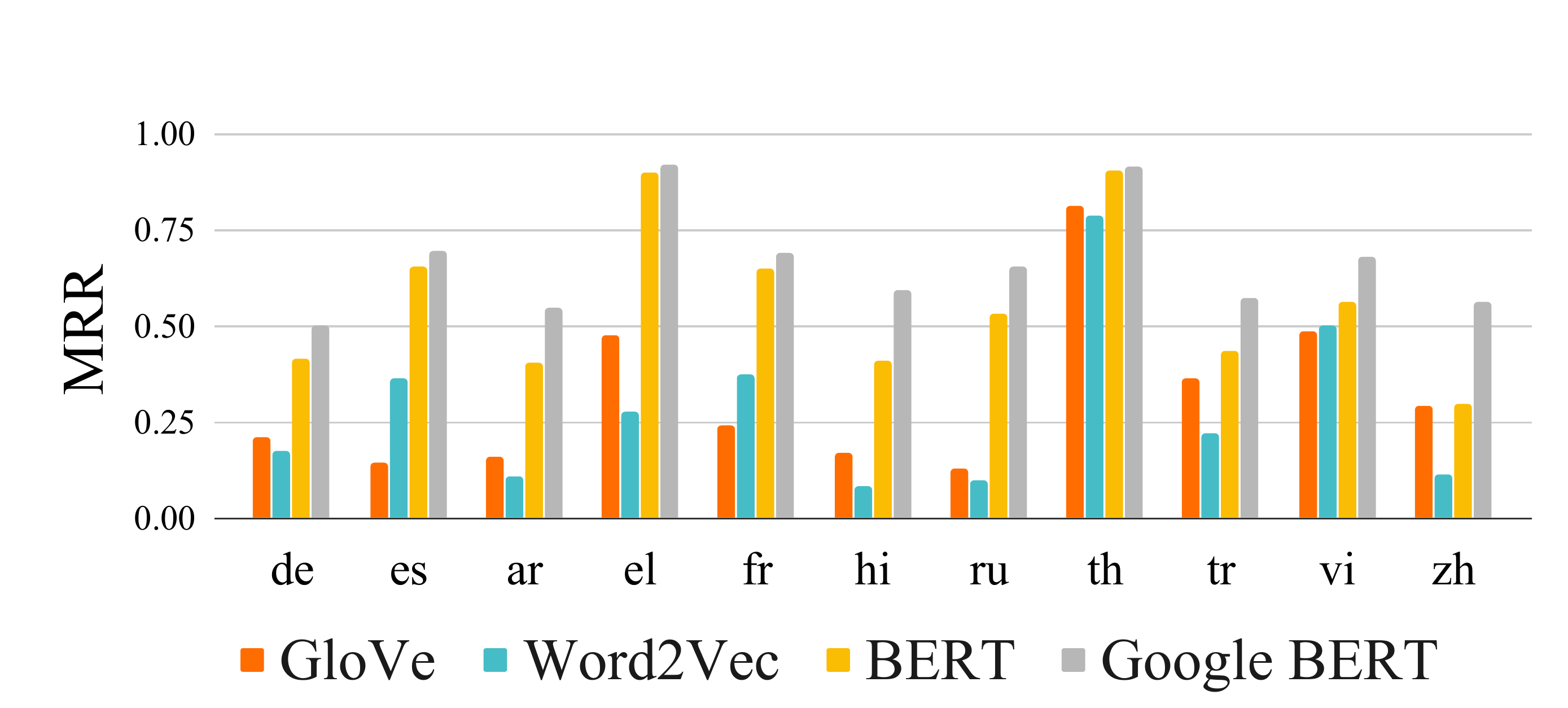}
 \subcaption{Pre-trained on 1000k sentences per language}
 \label{fig:3_3}
\end{subfigure}
\caption{Evaluating alignment with Word Retrieval. \footnotemark}
\vspace{-0.5cm}
\end{figure*}

\begin{figure*}[t!]
\centering 
\begin{subfigure}[b]{\columnwidth}
\includegraphics[width=\linewidth]{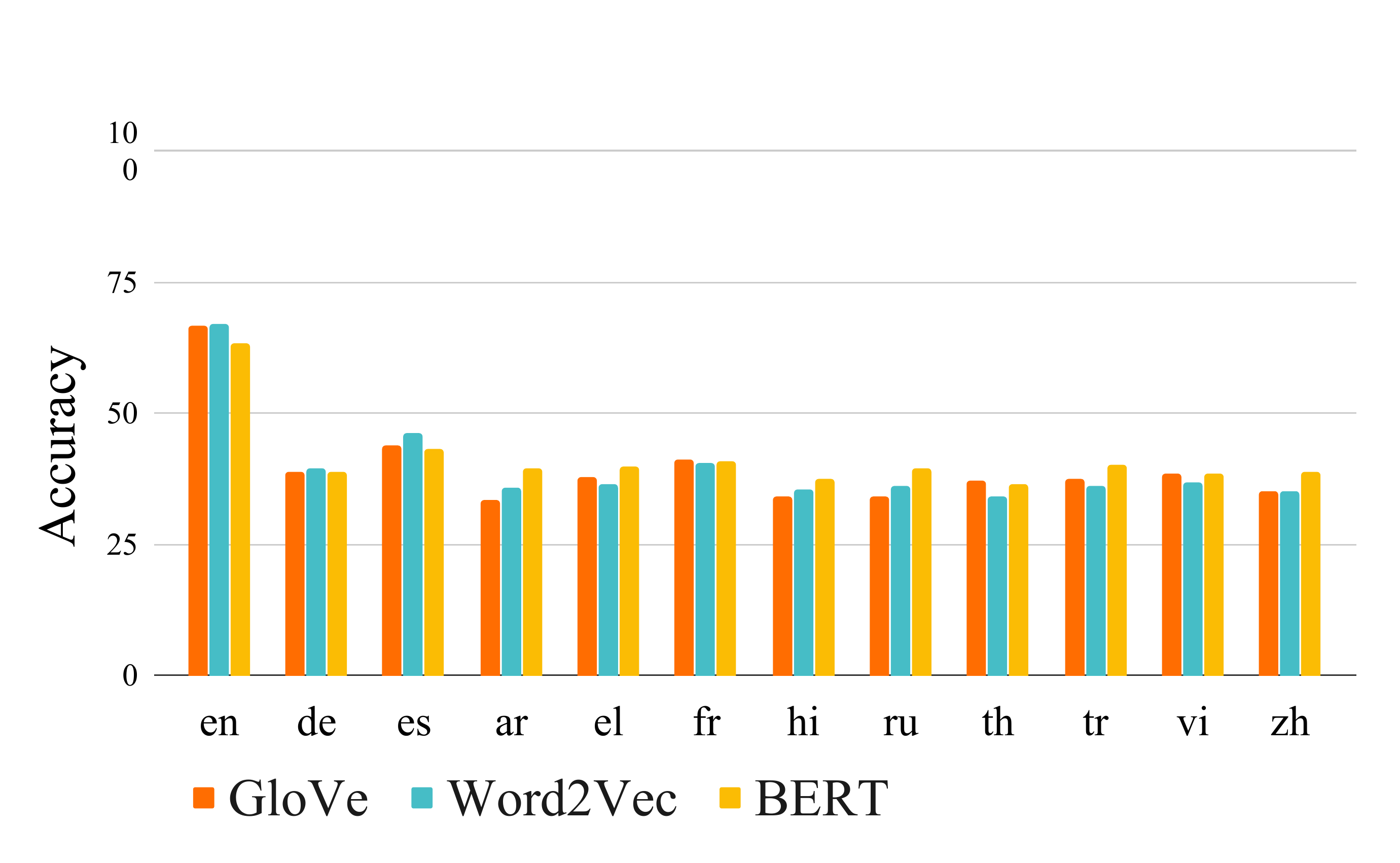}
 \subcaption{Pre-trained on 200k sentence per language}
 \label{fig:3_2}
\end{subfigure}
\hfill
\begin{subfigure}[b]{\columnwidth}
\centering 
\includegraphics[width=\linewidth]{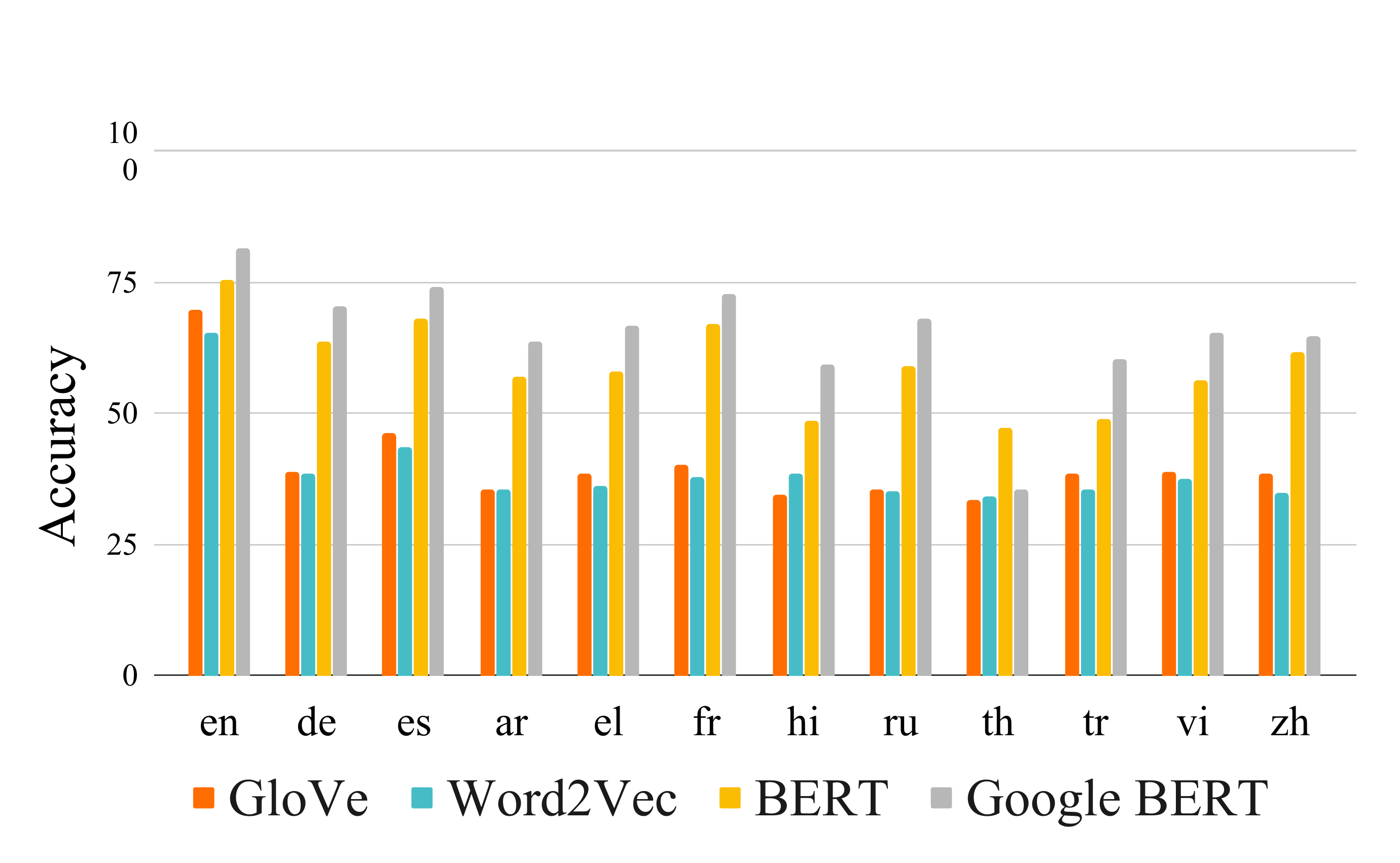}
 \subcaption{Pre-trained on 1000k sentence per language}
 \label{fig:3_4}
\end{subfigure}
\caption{Performance comparison on XNLI zero-shot cross-lingual transfer task \footnotemark[2]}
\vspace{-0.5cm}
\end{figure*}

We experimented with different amounts of data, {\it 200k} and {\it 1000k} sentences per language, to study the results of different models under different data sizes. 
The results are shown in Figure~\ref{fig:3_1},~\ref{fig:3_2},~\ref{fig:3_3} and~\ref{fig:3_4}.

\footnotetext{Google BERT is the m-BERT pre-trained by Google.}

\subsubsection{Small Pretraining Data}
\label{subsubsection:small}

Surprisingly, when pre-trained on small pretraining data (200k sentences per language), BERT didn't show its extraordinary cross-lingual ability, as shown in Figure~\ref{fig:3_1} and \ref{fig:3_2}. 
GloVe and Word2Vec achieved stronger cross-lingual alignment than BERT in terms of MRR score on word retrieval task on every language paired with English. 
Although BERT achieved better accuracies on XNLI zero-shot transfer on several languages, the margins were very small, and the overall performances were not better than GloVe and Word2Vec.  

This finding provides a further discussion of the literature. 
It has been found that the capacity of models is proportional to the ability to cross-lingual~\citet{Karth:20}\footnote{When the numbers of attention heads and total parameters were fixed, decreasing model depth decreased cross-lingual transfer performance; on the other hand, when the numbers of attention heads and depth were fixed, decreasing the number of network parameters degenerated performance, either.}. 
However, is it really the case that the bigger, the better? 
From the results here, the capacity of BERT is much bigger than GloVe and Word2Vec. Still, in the case of pretraining on the limited size of data, BERT didn't achieve superior performance as expected, suggesting that the relation of model capacity and cross-lingual ability may not be monotonic and the size of pretraining data also comes into play. 

\subsubsection{Big Pretraining Data}
\label{subsubsection:big}

When pre-trained on big pretraining data (1000k sentences per language), there was a dramatic turn as shown in Figure~\ref{fig:3_3} and~\ref{fig:3_4}. 
BERT achieved an overwhelmingly higher MRR score than other embeddings on every XX-En language pairs, showing that it did a much better job in aligning semantically similar words from different languages. 
Testing results on XNLI were also consistent with word retrieval task, BERT reached higher accuracies than GloVe and Word2Vec, demonstrating that it had the better cross-lingual ability.  

It was noticeable that the increase in pretraining data size largely improved the cross-lingual alignment and transferability of BERT, while it was not the same case for GloVe and BERT. And the bounding performance of Google BERT, which is the pretrained parameters released by Google, shows that there is still room for improvement if given even more pretraining data.


\begin{figure*}[t!]
\begin{subfigure}{\columnwidth}
  \includegraphics[width=\linewidth]{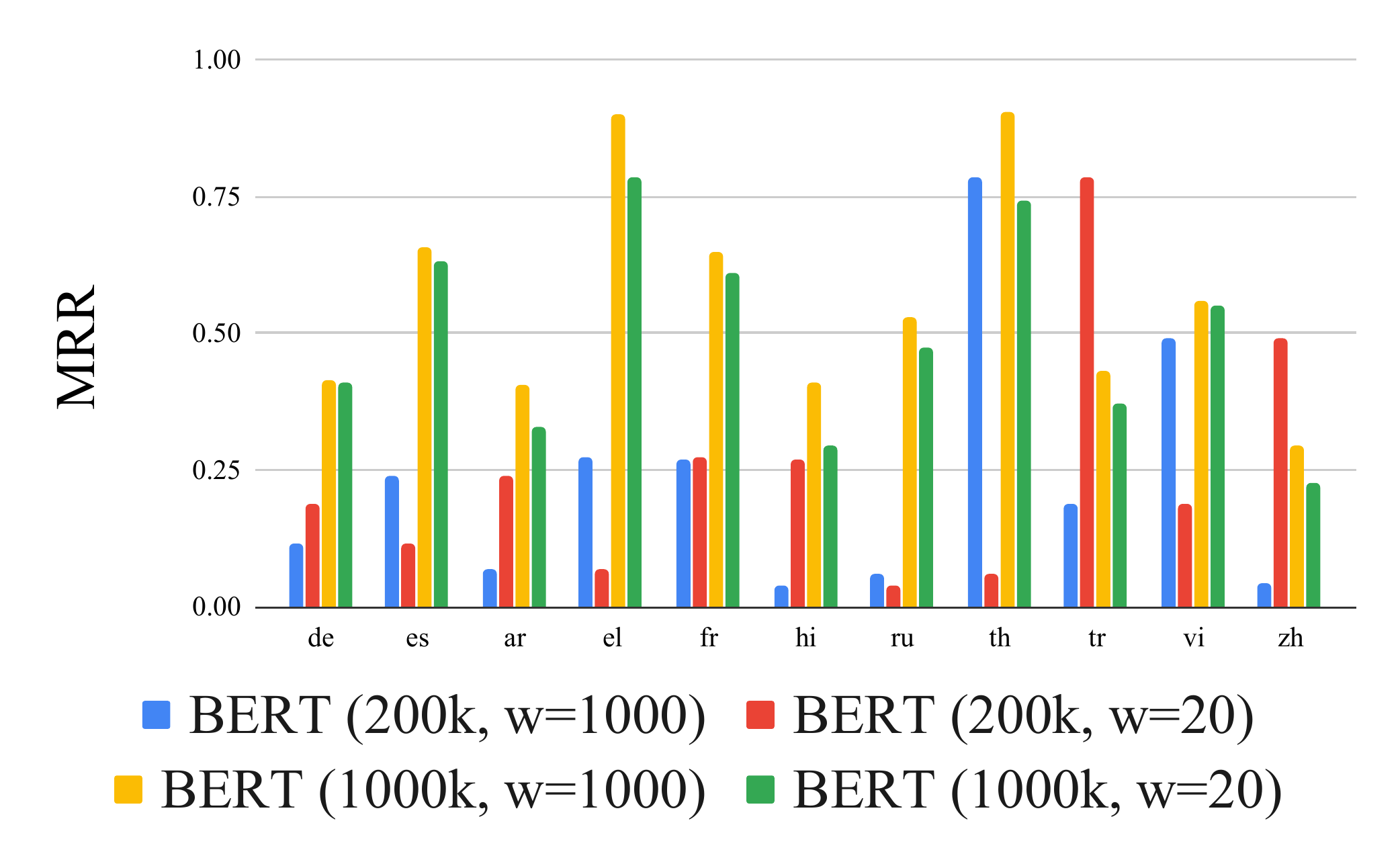}%
  \subcaption{Evaluating alignment with Word Retrival.}
\end{subfigure}
\hfill
\begin{subfigure}{\columnwidth}
  \includegraphics[width=\linewidth]{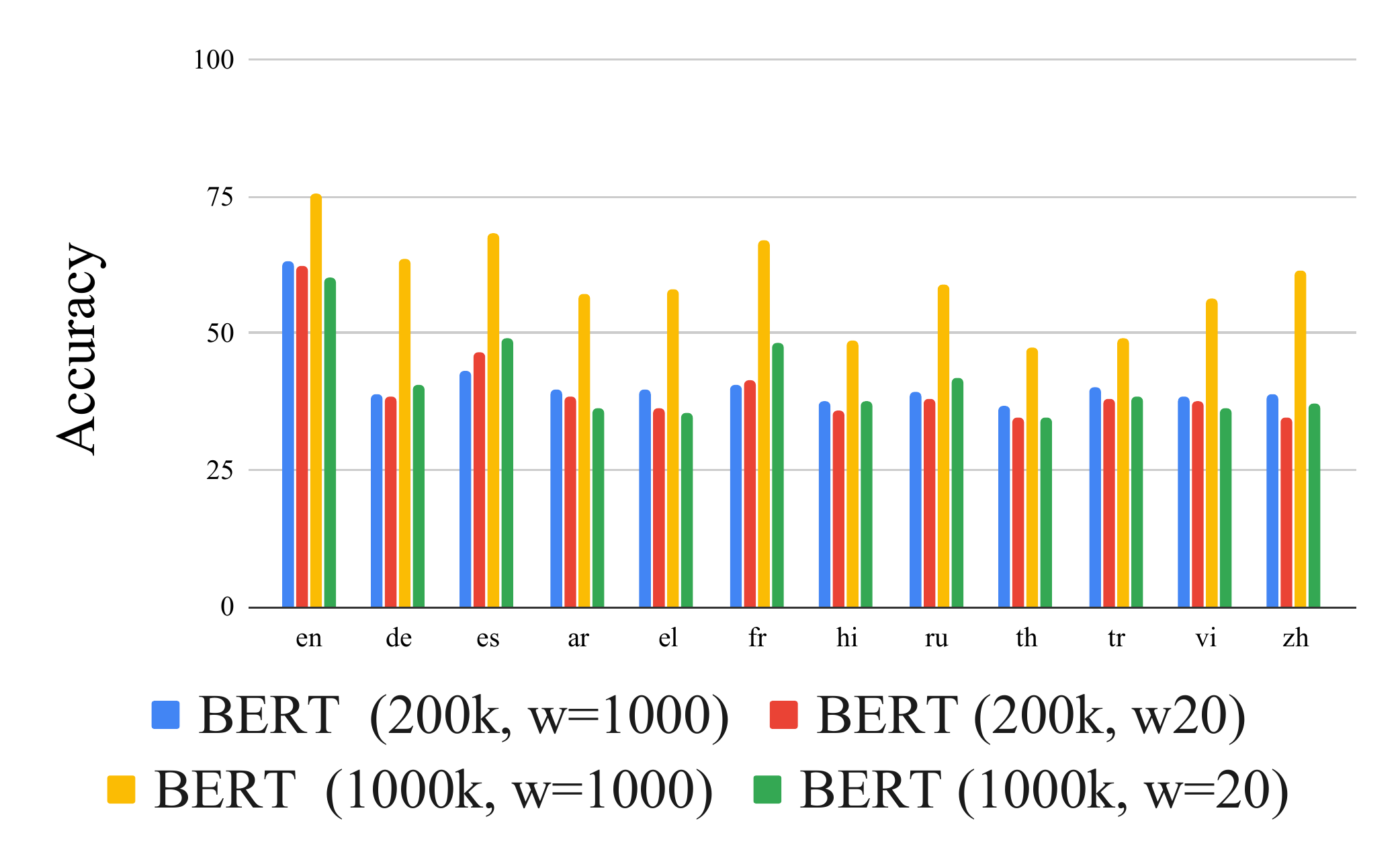}%
  \subcaption{Performance on XNLI zero-shot cross-lingual transfer task.}
\end{subfigure}
\caption{The Effect of Window Size $w$.}
\label{fig:3_5}
\vspace{-0.5cm}
\end{figure*}

\subsection{Breaking Down Long Dependency}

We noticed that in the literature, the typical co-occurrence window size of non-contextualized embeddings, like GloVe and Word2Vec, are often limited to 5$\sim$30 tokens, but BERT could attend to hundreds of tokens, which means that BERT could learn from longer dependency and richer co-occurrence statistics. 
Does the power of m-BERT come from a larger window size? 
We experimented with a smaller window size to find out if longer dependency is also necessary for learning cross-lingual structures. 
We directly sliced sentences in original pretraining data into smaller segments, limiting input length to 20 tokens for each example. And then we evaluated embeddings pretrained on these segments for cross-lingual ability\footnote{Limiting the number of tokens attended by attention heads may not work because the information from far tokens could still flow through layers and be collected at deeper layers.}.

The results of different window sizes are shown in Figure~\ref{fig:3_5}. In the case of big pretraining data (1000k), pretraining BERT with shortened inputs drastically hurt the cross-lingual ability of BERT, indicated by lower MRR score on word retrieval task compared to BERT pre-trained on normal-lengthed data. It should be noticed that the total number of tokens in pretraining data stayed unchanged.

However, in the case of small pretraining data (200k), pretraining BERT with shortened inputs yielded to better cross-lingual alignment on several languages, suggesting that breaking down long dependency helps BERT to even learn better cross-lingual alignment when only limited data is available.  

\begin{figure}[t!]
\centering 
\includegraphics[width=\linewidth]{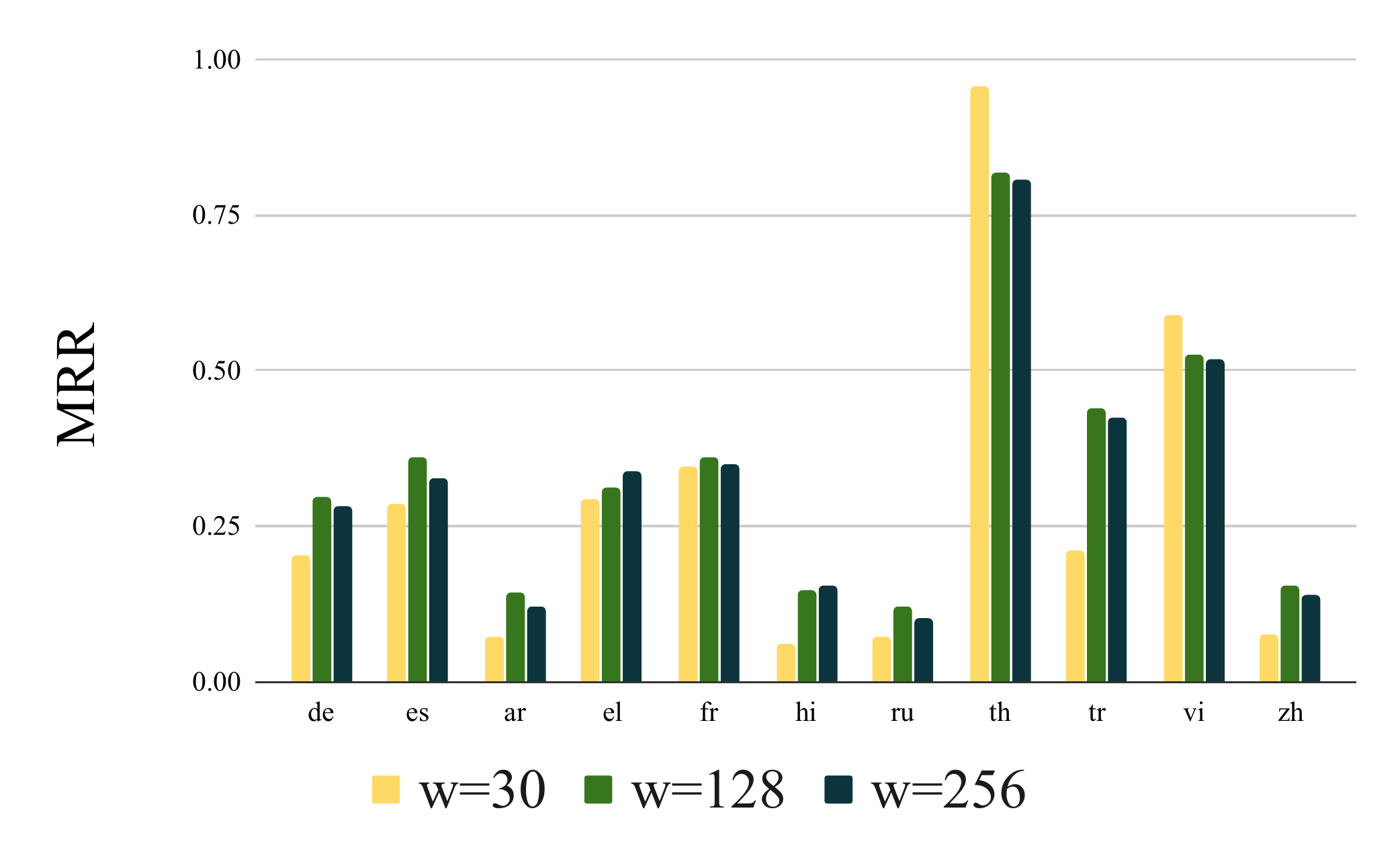}
 \caption{The Effect of Window Size $w$ on Word2Vec.}
 \label{fig:3_7}
 \vspace{-0.25cm}
\end{figure}

We further checked the case of Word2Vec, varying window size to 128 and 256 on small pretraining data. The alignment evaluation via MRR is shown in Figure~\ref{fig:3_7}. In most of the languages, increasing window size was not always beneficial, suggesting there may be a bottleneck from model capacity.

Considering the above observations, we hypothesized that cross-lingual ability of m-BERT is learned not only from local co-occurrence relations but also from co-occurrence relations of global scope, with a larger amount of data and model capacity required.

\subsection{Part-of-speech v.s. Cross-lingual}
At last, to gain insight into where alignment happens and which type of tokens are aligned best in our pretrained BERT (1000k), we analyzed the MRR score by part-of-speech (POS) tag obtained from the OntoNotes Release 5.0~\footnote{https://catalog.ldc.upenn.edu/LDC2013T19} dataset. 
We simply associated each token in m-BERT vocabulary with its most common POS label from OntoNotes annotations and calculated MRR for each class of POS tag. 
For simple comparison, we further grouped all the part-of-speech tags into closed-class and open-class two classes, as shown in table~\ref{tab:mrr_by_pos}, where fixed sets of words serving grammatical functions fall into closed-class and lexical words (e.g. noun, verb, adjective and adverb) fall into open-class. 
Different from \citet{Cao:20}, we showed that our BERT has higher alignment score for open-class versus closed-class categories. And among open-class of POS tags, adjectives are better aligned, on top of nouns, adverbs and verbs.        
\begin{table}[t!]
\centering
\caption{MRR by part-of-speech tag}
\label{tab:mrr_by_pos}
\footnotesize
\setlength\tabcolsep{3pt}
\begin{tabular}{l|cccccccc|c}
\toprule
\bf{MRR} & de & es & ar & fr & ru & tr & avg \\
\midrule
Closed & 0.392	& 0.558	& 0.310	& 0.600	& 0.403	& 0.491 & 0.459 \\
Open & \bf{0.419}	& \bf{0.577}	& \bf{0.387}	& \bf{0.647}	& \bf{0.493}	& \bf{0.542} & \bf{0.511} \\
\midrule
Noun	&0.466	&\bf{0.575}	&0.298	&0.623	&0.385	&0.534 &0.480 \\
Verb	&0.292	&0.513	&0.334	&0.512	&0.406	&0.306 &0.394 \\
Adv. &0.374	&0.507	&\bf{0.406}	&0.505	&0.484	&0.361 &0.440 \\
Adj. &\bf{0.482}	&0.574	&0.171	&\bf{0.625}	&\bf{0.529}	&\bf{0.567} &\bf{0.491} \\

\bottomrule
\end{tabular}
\vspace{-0.25cm}
\end{table}

\section{Conclusion}
In this paper, we find out that the cross-lingual ability of m-BERT has been learned from longer dependency (hundreds of tokens) instead of local co-occurrence information, and a massive amount of data is necessary. 
We also find that non-contextualized word embeddings cannot have the same cross-lingual ability even with the same amount of data and modeling the same length of dependency.


\bibliography{anthology,emnlp2020}
\bibliographystyle{acl_natbib}

\end{document}